\documentclass[conference]{IEEEtran}
\IEEEoverridecommandlockouts
\usepackage{cite}
\usepackage{amsmath,amssymb,amsfonts}
\usepackage{algorithmic}
\usepackage{afterpage}
\usepackage{graphicx}
\usepackage{placeins}
\usepackage{stfloats}
\usepackage{float}
\usepackage{textcomp}
\usepackage{xcolor}
\usepackage{multirow}
\def\BibTeX{{\rm B\kern-.05em{\sc i\kern-.025em b}\kern-.08em
    T\kern-.1667em\lower.7ex\hbox{E}\kern-.125emX}}

\begin{document}

\title{From Natural Language to SQL: Review of LLM-based Text-to-SQL Systems}

\author{
    \IEEEauthorblockN{Ali Mohammadjafari\IEEEauthorrefmark{1}, Anthony S. Maida\IEEEauthorrefmark{2}, Raju Gottumukkala\IEEEauthorrefmark{2}}
    \IEEEauthorblockA{\IEEEauthorrefmark{1}PhD Student, School of Computing and Informatics, University of Louisiana at Lafayette, LA, USA\\
    \IEEEauthorrefmark{2}Associate Professor, School of Computing and Informatics, University of Louisiana at Lafayette, LA, USA\\
    \IEEEauthorrefmark{2}Associate Professor, School of Mechanical Engineering, University of Louisiana at Lafayette, LA, USA\\
    \{Ali.mohammadjafari1, maida, raju\}@louisiana.edu}
}

\maketitle

\begin{abstract}
LLMs when used with Retrieval Augmented Generation (RAG), are greatly improving the SOTA of translating natural language queries to structured and correct SQL. Unlike previous reviews, this survey provides a comprehensive study of the evolution of LLM-based text-to-SQL systems, from early rule-based models to advanced LLM approaches that use (RAG) systems. We discuss benchmarks, evaluation methods, and evaluation metrics. Also, we uniquely study the use of Graph RAGs for better contextual accuracy and schema linking in these systems. Finally, we highlight key challenges such as computational efficiency, model robustness, and data privacy toward improvements of LLM-based text-to-SQL systems.
\end{abstract}


\section{Introduction}

\subsection{Overview of the Text-to-SQ task}

Organizations increasingly rely on relational databases to manage and analyze vast amounts of structured information. These databases are critical parts of many modern systems, from business intelligence to customer relationship management. As the volume of data increases, the need to query, extract, and make sense of this information also increases. However, querying databases often requires the use of Structured Query Language (SQL) which is a technical skill. There is a gap between users who need access to data and the specialized knowledge required to retrieve it \cite{kanburouglu2024text}. Text-to-SQL parsing in natural language processing (NLP) bridges this gap.

LLMs make the task of translating natural language (NL) queries given by a non-technical user to precise SQL easier. This ability is important when the complexity of data increases and makes manual data exploration impractical and inefficient \cite{dong2023c3}. For example, consider a relational database that contains information about gas stations in Louisiana, with columns such as {\footnotesize \texttt{GasStationID}}, {\footnotesize \texttt{PARISH}}, {\footnotesize \texttt{NEED}}, {\footnotesize \texttt{STATION NAME}}, {\footnotesize \texttt{CITY}}, and {\footnotesize \texttt{WORKING\_GAS\_STATIONS\_5\_MILES}} among others. Suppose we have a NL query: \emph{\textquotedblleft Where can I find a gas station with power less than 2 miles from the University?\textquotedblright} A text-to-SQL system would analyze this query, understand the user’s intent, and automatically generate the corresponding SQL query to extract the correct information from the database. In this case, the final SQL query might look something like:

{\footnotesize
\begin{verbatim}
SELECT STATION_NAME, location 
FROM gas_stations
WHERE fuel_available = 'Yes' 
  AND distance < 2 
  AND ST_Distance_Sphere(Point(long, lat), 
  Point(University_Long, 
  University_Lat)) < 2;
\end{verbatim}}

This SQL query retrieves all gas stations within a 2-mile radius of the University that have power. The text-to-SQL system enables the user to extract this specific information without needing to write the SQL query thus making complex data more accessible.

Stack Overflow shows that 51.52\% of professional developers use SQL in their works. However, SQL is often a barrier for non-technical users because of its technical nature. This leaves a gap between the vast stores of data housed in relational databases and the ability of many users to access related data efficiently \cite{hong2024next}.

 However, building reliable text-to-SQL systems is highly challenging. The complexity arises from several factors: 

\begin{itemize}

 \item \textbf{1. Ambiguity and Linguistic Complexity:}
 
NL queries often include complex structures such as nested clauses, pronouns, or vague terms. A query might contain phrases like ``all gas stations in the area,'' where ``area'' could refer to a specific region not directly mentioned.

 \item \textbf{2. Schema Understanding:}
 
To generate accurate SQL queries, text-to-SQL systems must understand the database schema. In realistic applications, database schemas are often complex and can vary greatly between domains, making it challenging to map the NL query correctly to the database schema \cite{yu2019cosql}.

 \item \textbf{3. Rare and Complex SQL Operations:}
 
Some queries involve uncommon SQL operations, such as nested sub-queries, joins across multiple tables, or window functions. These operations can be difficult for text-to-SQL models to generate, especially if they are not frequently seen during training \cite{hong2024next}.

\textbf{4. Cross-Domain Generalization:}

Text-to-SQL models trained on one domain (e.g., customer service databases) may not generalize well to other domains (e.g., healthcare). Differences in schema structures, terminology, and query patterns make it difficult for models to perform consistently across a wide variety of databases \cite{gan2021exploring}.

\end{itemize}

Addressing these challenges is essential for the effectiveness of text-to-SQL systems. As databases become more complex, higher accuracy of text-to-SQL models is also needed. 

\subsection{Introducing Retrieval Augmented Generation as a Solution}
As already stated, despite advances in existing text-to-SQL systems, they still face limitations in schema understanding, handling complex queries, and generalizing across domains. Retrieval Augmented Generation (RAG) has emerged as a promising framework to address these challenges. RAG systems combine two key components:

\begin{itemize}

\item  A Retrieval Module that dynamically fetches relevant schema details, SQL query template, or domain-specific knowledge from structured and/or unstructured sources like documents \cite{chen2024benchmarking}. 

\item  A Generative Module that produces text-based output, such as SQL queries or direct answers by adding the retrieved context into the generation process. 

\end{itemize}

In this way RAG system instead of using the general knowledge of LLMs, and guessing the answers, use related external knowledge to generate corresponding answers. RAGs facilitate text-to-SQL tasks in two ways. 

\begin{itemize}

\item  Increasing SQL generation means that RAG retrieves schema-specific information, query examples, or templates to improve SQL query generation for complex or ambiguous queries in natural language \cite{orrenius2024enhancing}. 

\item  Bypassing SQL Generation means that instead of generating SQL, RAG can directly retrieve answers from data sources.

\end{itemize}

This dual capability makes RAG a powerful approach to address the limitations of text-to-SQL systems. By dynamically retrieving context and integrating it into the response generation process, RAG improves schema understanding, resolves linguistic ambiguities, and adapts to new domains without extensive retraining.

\subsection{Contributions of this Survey}
This survey provides a comprehensive review of text-to-SQL systems and explores how Retrieval-Augmented Generation (RAG) can address their limitations. The key contributions are:

\begin{itemize} 

\item This survey presents a comprehensive review of text-to-SQL systems, including their challenges, datasets, benchmarks, and evaluation metrics. 

\item This survey identifies important limitations in existing text-to-SQL systems, such as schema understanding, linguistic ambiguity, domain generalization, and complex query handling.

\item This survey explains how RAG complements and enhances text-to-SQL systems, addressing key challenges with dynamic retrieval capabilities.

\item This survey introduces a taxonomy to classify text-to-SQL techniques, and illustrates how RAG fits into this landscape.
 
\item This survey explains how Graph RAG became the state of the art and can enhance the text-to-SQL systems' accuracy. 
\end{itemize}

\begin{figure*}[t]
    \centering
    \includegraphics[width=.7\textwidth]{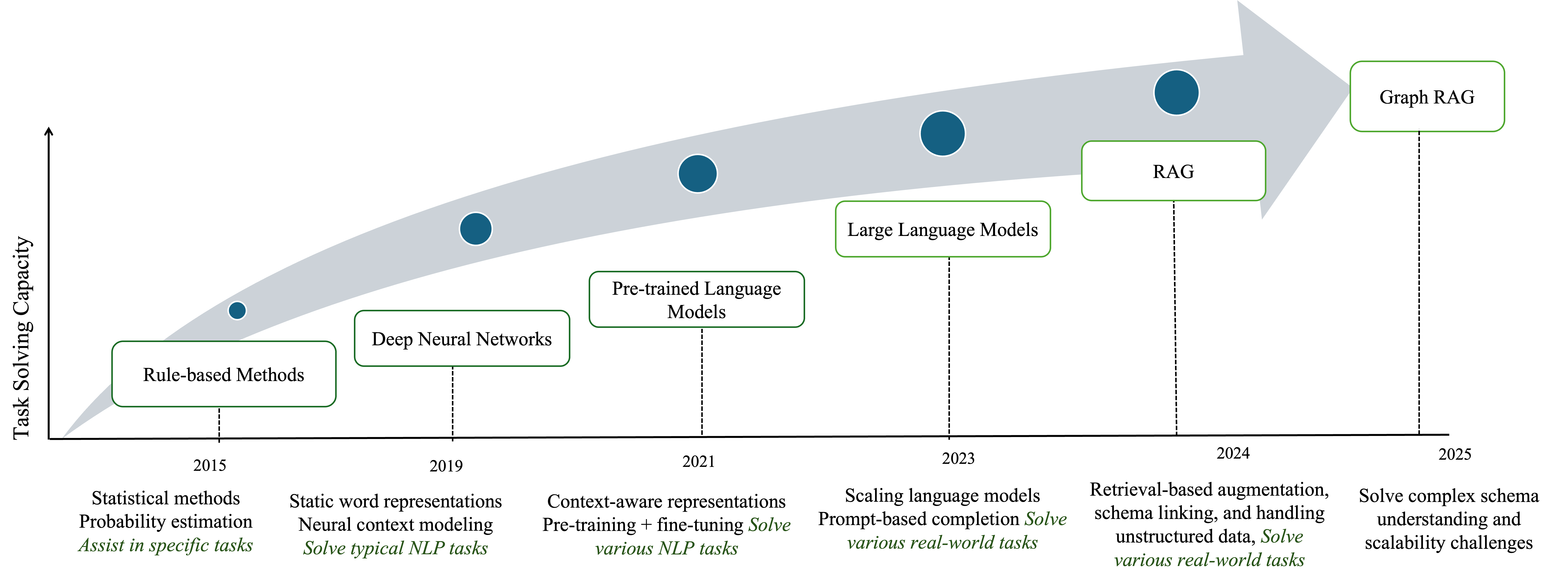}
    \caption{How text-to-SQL research has evolved over time, highlighting different implementation approaches. Each phase includes key techniques and notable works. The dates are approximate, based on when these key works were released, with a margin of error of about a year. The design is inspired by \cite{hong2024next}, \cite{zhao2023survey}.}
\label{fig:evolutionary}
\vspace{-.3 cm}
\end{figure*}

\section{Evolution of Text-to-SQL Systems in the Literature}

\subsection{Evolutionary Progression}

Figure~\ref{fig:evolutionary} shows the historical evolution of text-to-SQL systems. Initially they relied on rules, but now they use deep neural networks and powerful large language models (LLMs) \cite{hong2024next}. Each step in the evolution brought important innovations, making it easier for the newer models to understand and create SQL queries from everyday language.

\begin{figure*}[h]
    \centering
    \includegraphics[width=.85\textwidth]{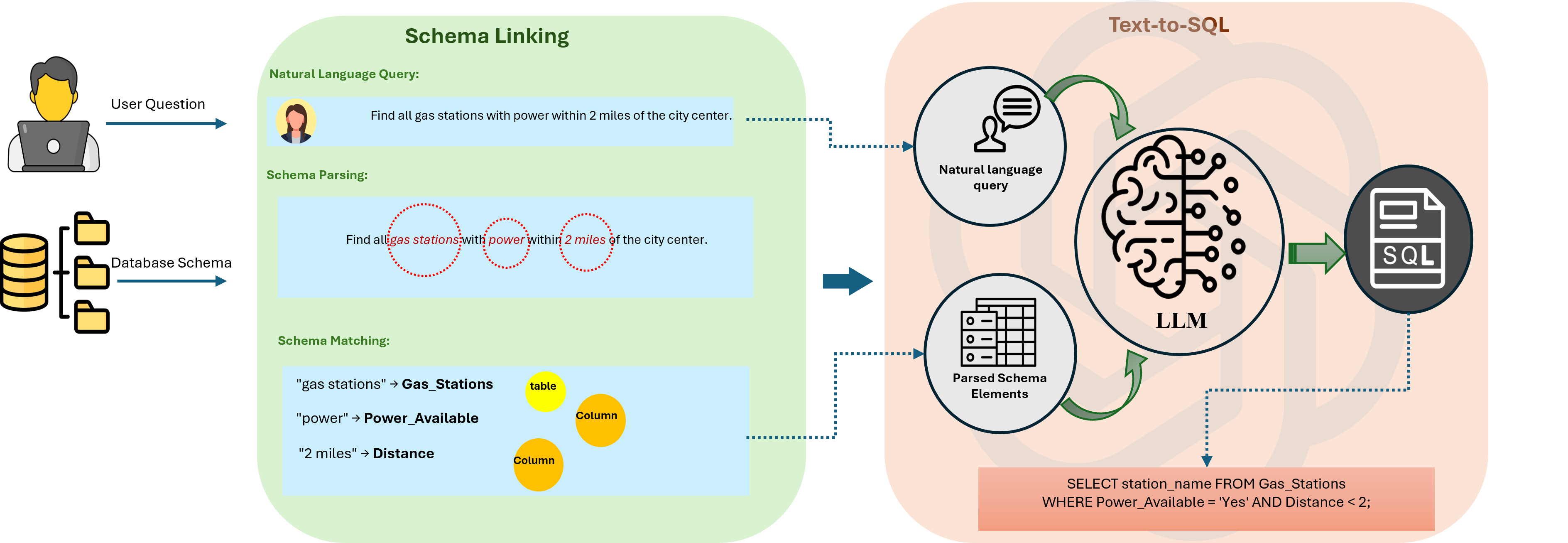}
    \caption{Illustrates the key stages of the traditional text-to-SQL process using Large Language Models (LLMs).}
\label{fig:arch}
\vspace{-.3 cm}
\end{figure*}

\begin{itemize}
     
\item \textbf{Rule-Based Approaches}

Early text-to-SQL systems used rule-based approaches. These systems used manually crafted grammar rules and heuristics to translate NL queries into SQL commands. These methods were limited in handling complex queries and diverse schemas \cite{mahmud2015rule}. The fixed rules made it hard to deal with the variety of human language, often causing errors when faced with tricky or unexpected query structures. Also these systems needed a lot of manual work to design and update the rules. This made it hard for them to adapt to new domains or database formats. Systems like LUNAR \cite{kang2013using} and NaLIX \cite{hammami2021automated} showed the potential of semantic parsing but required significant manual feature engineering, hindering scalability and adaptability \cite{kang2013using}, \cite{hammami2021automated}. Deep learning approaches have addressed many of these limitations by introducing models capable of learning complex patterns and representations directly from data \cite{el2024survey}. 

\item \textbf{ Deep Learning Approaches}

Deep learning greatly improved text-to-SQL systems by using large models to interpret NL and generate SQL queries. Models like Seq-2-SQL and SQLNet \cite{katsogiannis2021deep} used sequence-to-sequence architectures like LSTMs and transformers \cite{banitaba2024late}. These models introduced end-to-end, differentiable architectures that directly convert text to SQL, offering improvements over earlier systems and allowing the models to learn complex patterns and relationships within the data, leading to more accurate and efficient translation compared to earliear rule-based systems \cite{katsogiannis2021deep}, \cite{kumar2022deep}, \cite{naghshnejad2024graph}.

Transformer-based models like BERT \cite{li2024can} and TaBERT \cite{katsogiannis2023survey}, enhancedd understanding of both database schema and user intent. These models improved generalization to unseen databases by capturing dependencies between NL queries and underlying schemas. However challenges still remained such as handling nested queries, cross-domain generalization, and efficiently mapping ambiguous natural language to structured SQL still remained.  \cite{li2018deep}, \cite{katsogiannis2023survey}.

\item \textbf{Pre-trained Language Models}

 Pre-trained Language Models (PLMs) shifted the approach from task-specific supervised learning to a more generalized pre-training method followed by fine-tuning. Models like BERT \cite{li2024can} (Bidirectional Encoder Representations from Transformers) and GPT \cite{brown2020language} (Generative Pre-training Transformer) started this shift by using large-scale, unsupervised text datasets to pre-train models that could then be fine-tuned for specific tasks. The concept of PLMs, grounded in transfer learning, allowed models to acquire a deep understanding of NL through extensive pre-training on vast datasets, and this understanding transferred to many of tasks, like text generation, sentiment analysis, and question-answering \cite{tahir2024benchmarking}, \cite{wang2023pre}.

PLMs were better at capturing semantic relationships between user queries and database schema structures. PLMs such as TaBERT and BERT-SQL \cite{guo2019content} enabled the integration of both the NL query and the database schema into a unified representation that improved the system context. These models addressed several challenges in text-to-SQL systems, such as handling complex queries with multiple joins, nested queries, and cross-domain generalization. However, PLMs still had limitations in regards to their need for domain-specific fine-tuning and the difficulty in understanding complex database schemas without additional schema linking mechanisms \cite{10.1145/3605943}. 

\item \textbf{Large Language Models}

The advent of Large Language Models (LLMs) such as GPT-4 \cite{openai2023gpt4}, Codex \cite{chen2021codex}, and LLaMa \cite{touvron2023llama} has revolutionized NL processing. LLMs are a major advance over earlier machine learning methods by virtue their vast size and training on massive datasets to generate more accurate and comprehensive responses. These models show exceptional performance in tasks that require understanding and generating human-like text, often without needing additional fine-tuning \cite{rajkumar2022evaluating}. 

LLMs capture more complex relationships between NL queries and structured database schemas than the earlier models. Unlike previous pre-trained language models, which require significant task-specific fine-tuning and schema linking, LLMs can handle zero-shot and few-shot scenarios more effectively because of their large-scale pre-training and reasoning capabilities \cite{zhang2024benchmarking}. Studies show that models like Codex \cite{rajkumar2022evaluating} achieve high performance in generating SQL queries with minimal prompt engineering. However, challenges such as handling ambiguous queries and optimizing SQL statements for performance and correctness remain \cite{rajkumar2022evaluating}. 

As seen in Figure~\ref{fig:arch}, the architecture of LLM-based text-to-SQL systems can be broken down into several key phases: natural language understanding, schema comprehension, SQL generation, and SQL execution. Each step involves sophisticated techniques to ensure that user queries are accurately mapped to SQL, providing correct and meaningful results from database.

\item \textbf{RAG Systems in Text-to-SQL}

The evolution of text-to-SQL system has seen significant advancements, with the integration of RAG marking another step forward. By combining retrieval mechanisms with large-scale generative models, RAG systems address some of the persistent limitations in text-to-SQL tasks, particularly in schema understanding, handling complex queries, and domain generalization. 
\begin{itemize}
    \item \textbf{Dynamic Knowledge Retrieval:} 
     RAG systems utilize retrieval modules to fetch relevant schema information, table relationships, and example queries from structured or unstructured sources, such as relational databases or external documents. These retrieved elements are then integrated into the generative process, providing real-time context to improve SQL generation \cite{vichev2024ragsql}, \cite{orrenius2024enhancing}.

    \item \textbf{Enhanced Schema Linking:} 
    Unlike earlier models that relied heavily on pre-defined schema representations, RAG systems dynamically adapt to schema complexities. They align user queries with database schemas more effectively by retrieving schema-specific details during query processing, thus reducing errors caused by schema ambiguity \cite{zhao2024chat2data} \cite{zhang2023refsql}. 

    \item \textbf{Cross-Domain Generalization:}
    Traditional text-to-SQL systems often struggle to generalize across different database schemas or domains. RAG systems mitigate this challenge by leveraging domain-specific retrieval mechanisms, enabling seamless generalization without extensive fine-tuning. This makes RAG systems particularly effective for zero-shot or few-shot scenarios \cite{zhao2024chat2data}.

\end{itemize}
\item \textbf{Graph RAG System}
Graph RAG \cite{edge2024local} offers a structured and hierarchical approach to Retrieval-Augmented Generation (RAG), making it a promising solution for Text-to-SQL tasks. Unlike traditional semantic search methods that rely solely on text-based retrieval, Graph RAG constructs a knowledge graph from raw data, organizes schema elements into a hierarchical structure, and generates summaries for different database components. By leveraging these structured relationships, GraphRAG enhances schema understanding, improves retrieval accuracy, and enables more precise SQL query generation, making it particularly effective for complex database interactions.

\end{itemize}

\vspace{-5pt}

\subsection{LLM-based Text-to-SQL Architecture and RAG-Integrated Systems}

\begin{itemize}
    
\item \textbf{Natural Language Understanding:}

In traditional LLM-based text-to-SQL systems, the process begins with user input in the form of NL queries. LLMs first reprocess these input to understand the user's intent, identifying key components such as entities, conditions, and relations within the question.

\item \textbf{Schema Linking:}

Once the NL query is parsed, the system moves to schema linking, where the LLM maps the parsed components of the query to the corresponding tables, columns, and relationships in the database schema. For example, ``gas stations'' is linked to a table named \texttt{GasStations}, and ``power'' is matched with a column named \texttt{PowerAvailable}. This phase ensures that the system can correctly interpret the query.

\item \textbf{SQL Generation:}

After the query has been parsed and linked to the schema, the LLM generates an SQL query based o the established semantic relationships. This stage uses the model's understanding of SQL syntax and database logic to form a structured query that reflects the user's intent. The generated SQL is then validated and optimized for accuracy and performance. 

\item \textbf{SQL Execution and Output:}

The final SQL query is executed on the underlying database (such as SQLiteor MySQL) to retrieve the requested information. The results of the query are returned either in raw format or, in some systems, converted back into NL for easier interpretation by the user. Figure \ref{fig:arch} shows the flow from user input to the final SQL query. Each phase makes the text-to-SQL systems more accessible for non-technical users.
\end{itemize}

RAG improves the traditional text-to-SQL architecture by integrating dynamic retrieval mechanisms with LLMs to address challenges like schema complexity, ambiguous queries, and domain generalization. 

\begin{itemize}
\item \textbf{1. Improved Natural Language Understanding:}

RAG-to-SQL systems dynamically retrieve relevant external knowledge: 1. Schema metadata, example queries, and documentation are fetched from a vector database based on the input NL query. 2. If the query is ambiguous, RAG systems retrieve clarifying context, such as domain-specific examples or schema descriptions, to refine understanding.

\item \textbf{2. Schema Linking with Contextual Retrieval:}

In RAG-to-SQL systems, RAG retrieves detailed schema information, including table-column, relationships, foreign keys, and other data, and help more precise linking of query component. 

\item \textbf{3. Advance SQL Generation:}

The retrieved schema details, query examples, and metadata are integrated into the prompt, guiding the LLM to generate SQL queries that are both correct and semantically aligned with the database.

\item \textbf{4. Iterative SQL Execution and Error Feedback:}

RAG-to-SQL systems incorporate a feedback loop to address execution errors. SQL execution errors are detected during query execution. These errors are used to retrieve additional context from the vector database. The SQL query is re-generate and re-executed until the errors are resolved. 

\end{itemize}

\begin{figure*}[h]
    \centering
    \includegraphics[width=.85\textwidth]{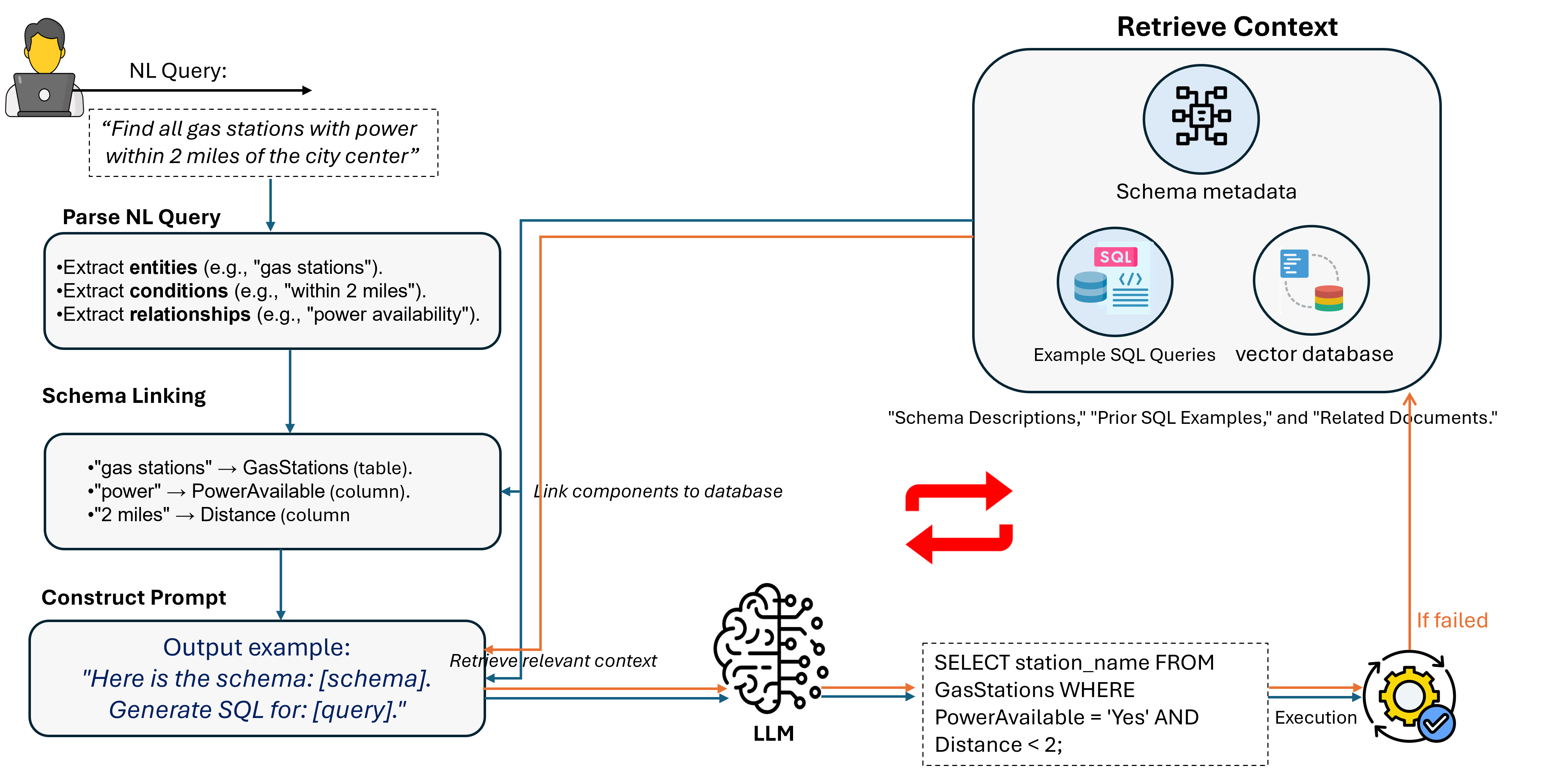}
    \caption{Illustrates the High-Level Workflow of RAG-based Text-to-SQL System (RAG-TO-SQL).}
\label{fig:ragarch}
\vspace{-.3 cm}
\end{figure*}

In Table ~\ref{table:high}, High-level comparison between the LLM-based Text-to-SQL System and RAG-Integrated Systems architectures is shown. And also in Fig.\ref{fig:ragarch} the high level workflow of RAG-to-SQL is illustrated.

\begin{table*}[]
\centering
\caption{Comparison between LLM-based Text-to-SQL System and RAG-Integrated Systems architectures}
\label{table:high}
\small
\begin{tabular}{|l|l|l|}
\hline
\textbf{Phase} & \textbf{Traditional LLM-Based Text-to-SQL} & \textbf{RAG-Integrated Text-to-SQL} \\ \hline
Natural Language Understanding & 
\begin{tabular}[l]{@{}l@{}}LLM parses user intent based on \\ pre-trained knowledge.\end{tabular} & 
\begin{tabular}[l]{@{}l@{}}RAG retrieves schema descriptions or \\ prior queries to enhance LLM’s understanding.\end{tabular} \\ \hline
Schema Linking & 
\begin{tabular}[l]{@{}l@{}}Relies on the LLM's schema understanding \\ and training data.\end{tabular} & 
\begin{tabular}[l]{@{}l@{}}Dynamically retrieves schema relationships \\ and metadata to improve linking accuracy.\end{tabular} \\ \hline
SQL Generation & 
\begin{tabular}[l]{@{}l@{}}Generates SQL based on in-context learning \\ or fine-tuning.\end{tabular} & 
\begin{tabular}[l]{@{}l@{}}Leverages retrieved examples/templates and \\ supports iterative refinement via feedback loops.\end{tabular} \\ \hline
SQL Execution & 
Executes SQL and retrieves raw results. & 
\begin{tabular}[l]{@{}l@{}}Refines SQL based on execution errors or \\ retrieves answers directly from sources.\end{tabular} \\ \hline
\end{tabular}
\end{table*}

\section{BENCHMARKS AND EVALUATION METHODS}

\begin{figure*}[h]
    \centering
    \includegraphics[width=.9\textwidth]{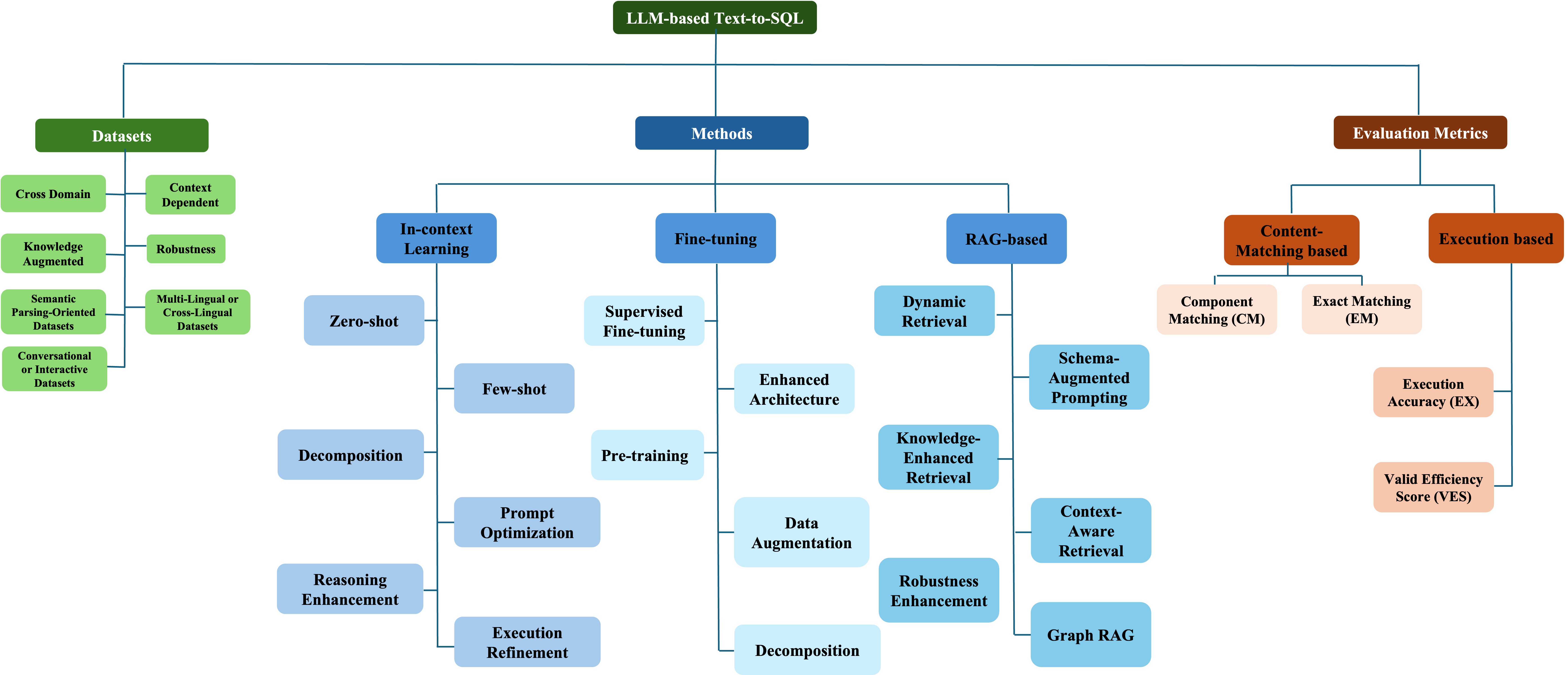}
    \caption{Taxonomy of research approaches in LLM-based text-to-SQL. The format is adapted from \cite{xu2023large}.}
\label{fig:taxonomy}
\vspace{-.3 cm}
\end{figure*}

\vspace{-5pt}
Evaluating LLM-based text-to-SQL systems is crucial for measuring how well they understand NL and generate accurate SQL queries. Researchers use a variety of datasets and benchmarks that test these models across different scenarios, both within specific domains and across multiple domains.

\begin{table*}[]
\centering
\caption{Comparison of Dataset Types in Text-to-SQL: Key Characteristics, Strengths, Challenges, and Examples \cite{zhong2017seq2sql}, \cite{yu2018spider}, \cite{yu2019cosql}, \cite{wang2020dusql}, \cite{li2024can}, \cite{yu2019sparc}, \cite{gan2021exploring}, \cite{gan2021towards}. \cite{deng2020structure}, \cite{goodman2019geoquery}, \cite{tur2010left}, \cite{zhang2024finsql}, \cite{orrenius2024enhancing}.}
\label{tab:Tdatasets}
\renewcommand{\arraystretch}{1.2}
\resizebox{\textwidth}{!}{
\begin{tabular}{|p{2.5cm}|p{3.5cm}|p{4cm}|p{3.5cm}|p{4cm}|p{4cm}|p{2.5cm}|}
\hline
\textbf{Category} & \textbf{Focus} & \textbf{Relation to RAG-to-SQL} & \textbf{Strength} & \textbf{Weaknesses} & \textbf{Challenges (Posed to Models)} & \textbf{Examples} \\ \hline

\centering Cross-Domain & 
\centering Tests generalization across diverse database schemas and domains. & 
RAG dynamically retrieves schema-specific mappings and examples, enabling better handling of unseen schemas. & 
1. Evaluates generalization across diverse schemas. \newline 2. Most common benchmark for real-world adaptability. & 
1. Limited focus on schema-specific nuances. \newline 2. May oversimplify domain-specific challenges. & 
1. Handling unseen schemas without extensive fine-tuning. \newline 2. Mapping ambiguous queries to schema elements. & 
1. Spider \newline 2. WikiSQL \\ \hline

\centering Knowledge-Augmented & 
\centering Requires external domain knowledge or context beyond the schema itself. & 
RAG retrieves unstructured knowledge (e.g., FAQs, documents) to provide additional context for SQL generation. & 
1. Tests integration of external domain knowledge. \newline 2. Simulates real-world domain-specific challenges. & 
1. Requires external knowledge sources, increasing complexity. \newline 2. High reliance on retrieval accuracy. & 
1. Linking domain-specific knowledge to schema elements. \newline 2. Avoiding hallucinations during SQL generation. & 
1. Spider-DK \newline 2. BIRD \\ \hline

\centering Context-Dependent & 
\centering Multi-turn queries requiring understanding of previous turns in a conversation. & 
RAG retrieves prior turns' context or intermediate query states to ensure continuity across dialogue. & 
1. Evaluates continuity and coherence across dialogue turns. \newline 2. Tests memory and contextual adaptation. & 
1. Limited datasets available for conversational SQL. \newline 2. Requires additional context tracking and retrieval logic. & 
1. Handling ambiguity in multi-turn interactions. \newline 2. Efficiently retrieving prior turn information. & 
1. CoSQL \newline 2. SParC \\ \hline

\centering Robustness & 
\centering Evaluates performance under adversarial or ambiguous inputs. & 
RAG retrieves schema clarifications, synonyms, or disambiguation examples to handle perturbations. & 
1. Highlights system resilience to perturbations. \newline 2. Tests error handling under noisy input conditions. & 
1. May overemphasize edge cases that are rare in real-world applications. \newline 2. Often lacks domain realism. & 
1. Resolving synonym ambiguities. \newline 2. Handling minimal or incomplete schema descriptions. & 
1. Spider-Syn \newline 2. Spider-Realistic \\ \hline

\centering Semantic Parsing-Oriented & 
\centering Tests baseline NL-to-SQL translation accuracy and schema linking. & 
RAG can retrieve schema metadata or query examples to improve schema linking and parsing accuracy. & 
1. Focuses on core NL-to-SQL translation abilities. \newline 2. Serves as a foundation for all other categories. & 
1. Limited to simple queries and fixed schemas. \newline 2. Doesn’t test domain generalization or robustness. & 
1. Achieving precise semantic parsing for varied linguistic structures. \newline 2. Avoiding overfitting to simple datasets. & 
1. GeoQuery \newline 2. ATIS \newline 3. WikiSQL (single-turn queries) \\ \hline

\centering Multi-Lingual or Cross-Lingual & 
\centering Evaluates performance across multiple languages. & 
RAG retrieves schema metadata and query examples in the same language as the input, improving multi-lingual adaptation. & 
1. Tests adaptability across languages. \newline 2. Highlights multi-lingual and cross-lingual generalization capabilities. & 
1. Limited availability of high-quality multi-lingual datasets. \newline 2. Heavily language-dependent retrieval models. & 
1. Mapping language-specific terms to schema elements. \newline 2. Handling multi-language inconsistencies in schema linking. & 
1. mSpider \newline 2. DuSQL \\ \hline

\centering Real-World Application & 
\centering Benchmarks derived from real-world industry use cases. & 
RAG retrieves domain-specific schema relationships or external documentation to support domain-specific SQL generation. & 
1. Evaluates systems on realistic, domain-specific queries. \newline 2. High practical relevance for real-world applications. & 
1. Often lacks diversity in schema structure. \newline 2. Domain-specific datasets may be inaccessible or proprietary. & 
1. Integrating domain-specific external knowledge. \newline 2. Handling highly specific schemas with unique structures. & 
1. Financial SQL \newline 2. Nibiru \newline 3. ClinicalDB \\ \hline

\centering Conversational or Interactive & 
\centering Designed for interactive dialogue settings with evolving context. & 
RAG retrieves historical interaction logs or previous conversational turns to ensure continuity and context-awareness in SQL queries. & 
1. Simulates real-world conversational settings. \newline 2. Evaluates dynamic context management in multi-turn queries. & 
1. High dependency on accurate retrieval of prior interactions. \newline 2. Computationally intensive for large conversations. & 
1. Managing long conversational histories. \newline 2. Handling ambiguous or implicit references across turns. & 
1. CoSQL \newline 2. SParC \\ \hline

\end{tabular}
}
\end{table*}

\subsection{Types of Datasets used in Benchmarks}
Text-to-SQL research has made rapid progress with the help of many benchmark datasets, each contributing unique challenges for model development. These datasets are categorized into four types based on their characteristics: cross-domain, knowledge-augmented, context-dependent, and robustness. In Table~\ref{tab:datasets}, we categorized most well-known datasets according to these criteria. 

\begin{itemize} 

\item \textbf{1. Cross-domain Datasets}

Datasets like WikiSQL \cite{zhong2017seq2sql}, Spider \cite{yu2018spider}, and KaggleDBA \cite{lee2021kaggledbqa} focus on evaluating the generalization capabilities of models across multiple databases from different domains. These datasets test whether models can generate accurate SQL queries for databases they have not seen during training \cite{yu2018spider}. 

In RAG-TO-SQL systems, RAG has better performance in cross-domain settings where schemas differ significantly across database. Retrieval modules dynamically fetch schema details or examples from diverse domains, and help to improve robust generalization.  

\item \textbf{2. Knowledge-Augmented Datasets}
Datasets such as SQUALL \cite{shi2020potential} and BIRD \cite{li2024can} use external knowledge to improve the semantic of SQL generation. These datasets aim to enhance the model's comprehension by augmenting the schema with additional context, allowing for more accurate and better SQL generation \cite{li2024can}. Spider-DK \cite{gan2021exploring} adds domain knowledge requirements to the spider dataset. RAG systems can retrieve external documents or unstructured knowledge to handle questions needing additional context.

\item \textbf{3. Context-Dependent Datasets}
Datasets like CoSQL\cite{yu2019cosql} and SParc\cite{yu2019sparc} emphasize the conversational nature of querying databases, where previous interactions influence current queries. These datasets challenge models to maintain context throughout multi-turn interactions, making them essential for developing systems that can handle complex, dialog-driven database queries \cite{yu2019sparc}.

\item \textbf{4. Robustness Datasets}
A dataset like ADVETA \cite{pi2022towards} tests the robustness of text-to-SQL systems, by introducing adversarial table perturbations. This method tests if models are capable of handling unexpected changes in database schema or table structure, thereby assessing their adaptability to real-world scenarios\cite{pi2022towards}. The Table~\ref{tab:Tdatasets} datasets push the boundaries of text-to-SQL research, providing challenges across various dimensions, from domain generalization to contextual understanding and system robustness. Retrieval modules fetch clarifying schema descriptions or mappings, and helps LLMs resolve ambiguities introduced in the dataset.

\item \textbf{5. Semantic Parsing-Oriented Datasets}

These datasets designed for evaluating the semantic parsing capabilities of models, where precise mappings from natural language to SQL are tested \cite{goodman2019geoquery}. 

\item \textbf{6. Multi-Lingual or Cross-Lingual Datasets}

Test model performance across multiple languages, requiring the system to map non-English queries to SQL. etrieval modules in RAG systems can fetch schema mappings or query examples in the specific language of the input, enhancing multilingual performance \cite{wang2020dusql}.

\item \textbf{7. Conversational or Interactive Datasets}

These datasets designed for conversation settings, where context from previous turns must be considered \cite{yu2019cosql}. CoSQL extends Spider for conversational text-to-SQL tasks, and SParC \cite{yu2019sparc} is a dataset for conversational SQL queries with dependencies between turns. RAG systems can retrieve prior turns’ context dynamically, ensuring continuity across dialogue turns.
\end{itemize}
\begin{figure*}
    \centering
     \includegraphics[width=1\textwidth]{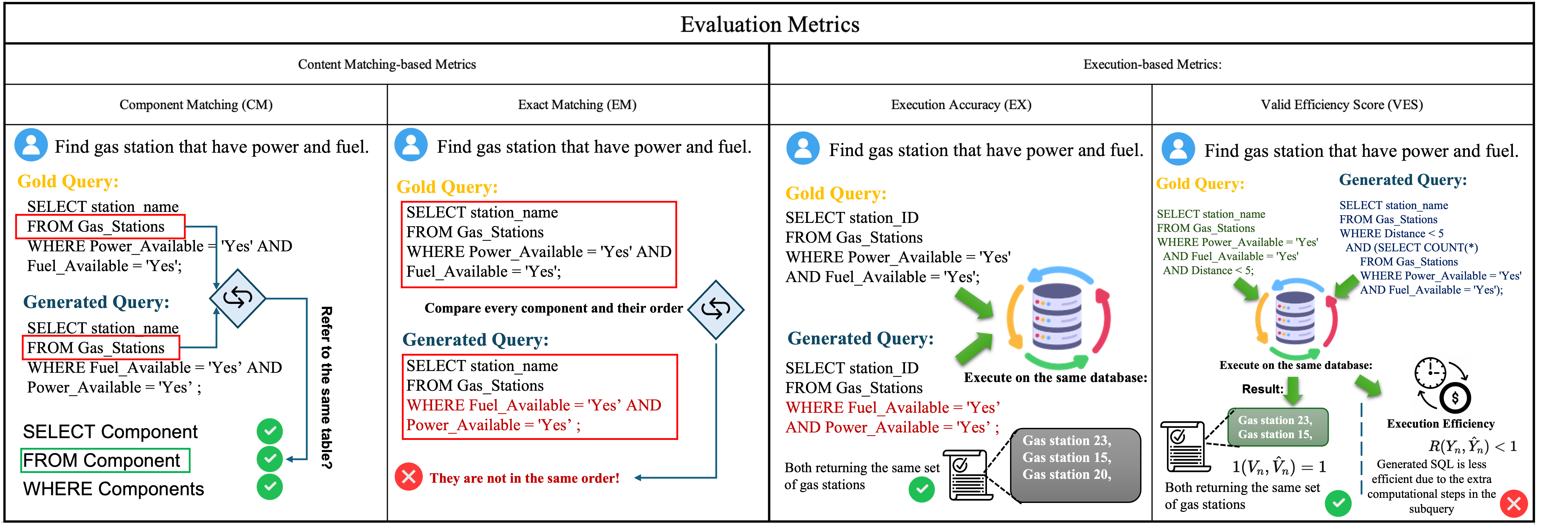}
    \caption{Four main evaluation metrics falling into two categories.}
\label{fig:metrics}
\vspace{-.3cm}
\end{figure*}

\subsection{Evaluation Metrics Used in Benchmarks}
Benchmarks for Text-to-SQL systems use metrics that capture both the correctness and efficiency of an SQL query. These metrics test that systems not only produce accurate SQL queries, but also perform efficiently in realistic database environments. Figure~\ref{fig:metrics} shows evaluation metrics that are fall into two categories: Content Matching-based Metrics and Execution-based Metrics.

\subsubsection{Content Matching-based Metrics}
Content matching-based metrics focus on how closely the structure of the generated SQL query matches the gold (or reference) query. This type of evaluation is needed for ensuring that the model follows the correct SQL syntax and structure, even if it might not always produce the most optimized SQL query.

\begin{itemize}  

\item \textbf{1. Component Matching (CM):} 

This metric evaluates each component of the SQL query ( such as SELECT, FROM, WHERE) individually. Even if the components appear in a different order than in the gold query, they are considered correct as long as they correspond to the expected components. This allows for flexibility in the query structure while ensuring the essential parts of the SQL query are present and accurate\cite{yu2018spider}.

\item  \textbf{2. Exact Matching (EM):}  

Exact matching is stricter, requiring the generated SQL query to match the gold query exactly in terms of both structure and order. Every element including the sequence of components, must be identical to the gold query. The disadvantage of this metric is it can penalize queries that are functionally correct but structured differently\cite{yu2018spider}. 

\end{itemize}

\subsubsection{Execution-based Metrics}
 Focus on the actual performance of the generated SQL query when run on a database. These metrics assess whether the queries not only follow the correct structure but also return the correct results and run efficiently in realistic scenarios. 

\begin{itemize}
     
\item \textbf{1. Execution Accuracy (EX):} 

Checks whether the generated SQL query, when executed on the database, returns the same result as the gold query. Execution accuracy focuses on the correctness of the result, regardless of how the query is structured \cite{yu2018spider}. 

\item \textbf{2. Valid Efficiency Score (VES):}

The Valid Efficiency Score measures the computational efficiency of the generated SQL query compared to the gold query. While a query might return the correct result, it could still be inefficient, requiring unnecessary computational resources. VES penalizes queries that introduce extra complexity, such as redundant sub-queries or unnecessary joins, even if the results match.\cite{yu2019cosql}, \cite{yu2019cosql}.
\end{itemize}

\subsection{Methods}
LLM-based text-to-SQL systems fall into two main classes of methods: \textbf{In-context learning} and \textbf{Fine-tuning (FT)}. Each class use different strategies for training the model to generate SQL queries from NL inputs. These methods rely on using the model's pre-trained knowledge through prompts or fine-tuning with added task-specific data to improve query generation. 

\begin{table*}[h]
\centering
\renewcommand{\arraystretch}{1.2}
\caption{comparison table of State of the art in RAG-Based text-to-SQL systems vs some new traditional LLM-based text-to-SQL}
\label{ICL}
\small
\resizebox{\textwidth}{!}{
\begin{tabular}{|p{3cm}|p{2cm}|p{2.5cm}|p{3.5cm}|p{2.5cm}|p{3cm}|p{3.5cm}|p{3cm}|p{2cm}|}
\hline
\textbf{Method Name} & \textbf{LLM Used} & \textbf{Database Used} & \textbf{Metrics} & \textbf{Category} & \textbf{Sub-category} & \textbf{Novelty} & \textbf{Weakness} & \textbf{Reference} \\ \hline

\centering Chat2Data & \centering Not specified & \centering Not specified & Execution accuracy & RAG-based & Knowledge-Enhanced Retrieval, Decomposition & Uses vector databases to enhance prompt generation & Lacks LLM specification and detailed dataset usage & \cite{zhao2024chat2data} \\ \hline

\centering Sample-aware Prompting and Dynamic Revision Chain & \centering GPT-3.5 & \centering Not specified & Exact matching (EM), Execution accuracy & RAG-based & Dynamic Retrieval, Context-Aware Retrieval & Dynamic revision chain for fine-grained feedback & Increased computational cost & \cite{guo2023retrieval} \\ \hline

\centering RAG-based Text-to-SQL & \centering GPT-4o-mini & \centering PostgreSQL & Exact matching (EM), Execution accuracy & RAG-based & Knowledge-Enhanced Retrieval & Direct text-to-SQL query generation & Struggles with database normalization principles & \cite{syrja2024retrieval} \\ \hline

\centering RAG-enhanced LLMs & \centering GPT-3.5, GPT-4, Llama2 & \centering Not specified & Exact matching (EM), Efficiency & RAG-based & Knowledge-Enhanced Retrieval & Evaluation across LLMs with and without RAG integration & Increased query generation times & \cite{bartczak2024rag} \\ \hline

\centering TAG & \centering Not specified & \centering Custom datasets & Exact match, Accuracy & RAG-based & Knowledge-Enhanced Retrieval, Schema-Augmented Prompting & Unified approach combining RAG and Text2SQL paradigms & High computational costs for iterative data synthesis & \cite{biswal2024text2sql} \\ \hline

\centering Context-Aware Generation & \centering Not specified & \centering OutSystems data & Execution accuracy & RAG-based & Dynamic Retrieval, Context-Aware Retrieval & Handling large context sizes by reducing irrelevant data & Increased inference time despite improved accuracy & \cite{henriques2024augmenting} \\ \hline

\centering CRUSH4SQL & \centering GPT-3 & \centering SPIDER, BIRD, SocialDB & Recall, Accuracy & RAG-based & Schema-Augmented Prompting, Context-Aware Retrieval & Leverages hallucination as a mechanism for high-recall schema subsetting & Increased complexity and resource requirements for schema extraction & \cite{kothyari2023crush4sql} \\ \hline

\centering FATO-SQL & \centering Medium-scale LLMs & \centering Petrochemical data & Accuracy & RAG-based & Context-Aware Retrieval, Dynamic Retrieval & Four-stage process with two rounds of LLM calls to improve SQL generation & Scalability is limited by medium-scale LLM parameters & \cite{chen2024fato} \\ \hline

\centering Distyl-SQL & \centering Proprietary LLM & \centering Not specified & Execution accuracy & RAG-based & Hierarchical CTE, Knowledge Management & Incorporates pre-processing for hierarchical decomposition of SQL generation & High latency and needs adaptive knowledge enhancement & \cite{maamari2024end} \\ \hline

\centering TARGET & \centering Not specified & \centering Various & Accuracy, Recall & RAG-based & Table Retrieval & Benchmarking retrieval effectiveness for generative tasks over tabular data & High variability in retriever performance across datasets & \cite{ji2024target} \\ \hline

\centering XRICL & \centering Codex & \centering SQLite & Component Matching (CM), Execution accuracy & In-context & Zero-shot, Prompt Optimization & Translation as Chain-of-Thought prompt for non-English Text-to-SQL & Limited cross-lingual retrieval exemplars availability & \cite{shi2022xricl} \\ \hline

\centering RSL-SQL & \centering GPT-4o, DeepSeek & \centering Spider, BIRD & Execution accuracy & Fine-tuning & Enhanced Architecture, Data Augmentation & Bidirectional schema linking to mitigate schema pruning risks & Increased complexity and risks from multi-turn strategies & \cite{cao2024rsl} \\ \hline

\centering TCSR-SQL & \centering Not specified & \centering Custom benchmark & Execution accuracy, Exact-set-match & In-context & Reasoning Enhancement, Schema-Augmented Prompting & Self-retrieval and multi-round generation-execution-revision process & High costs for fuzzy testing with large datasets & \cite{xu2024tcsr} \\ \hline

\centering E-SQL & \centering Not specified & \centering BIRD & Execution accuracy & In-context & Question Decomposition, Execution Refinement & Direct schema linking and candidate predicate generation & Diminishing returns when used with advanced LLMs & \cite{alp2024sql} \\ \hline

\centering DataGpt-SQL & \centering DataGpt-SQL-7B & \centering Spider & Accuracy (EX, TS) & Fine-tuning & Data Augmentation, Decomposition & Uses preference datasets and self-refine mechanisms to mitigate LLM risks & Requires fine-tuning on large datasets, potentially increasing costs & \cite{wu2024datagpt} \\ \hline

\centering SEA-SQL & \centering GPT-3.5 & \centering Spider, BIRD & Execution accuracy & In-context & Reasoning Enhancement, Adaptive Refinement & Adaptive bias elimination and dynamic execution adjustment & Limited efficiency when using complex schema structures & \cite{li2024sea} \\ \hline

\centering TA-SQL & \centering GPT-4 & \centering BIRD, Spider & Execution accuracy & In-context & Task Alignment, Schema Linking, Logical Synthesis & Task alignment to mitigate hallucinations, providing robustness in SQL generation & Dependence on task alignment makes implementation more complex & \cite{qu2024before} \\ \hline

\centering Interactive-T2S & \centering GPT-4, ChatGPT & \centering BIRD, Spider-Dev & Execution accuracy & In-context & Multi-Turn Interaction, Prompt Optimization & Stepwise generation with four tools for proactive information retrieval & Lack of scalability for wide tables & \cite{xiong2024interactive} \\ \hline

\centering SQLfuse & \centering Open-source LLMs & \centering Spider & Execution accuracy & Fine-tuning & Enhanced Architecture, SQL Critic & SQL generation with a critic module for continuous improvement & Complex module integration requires high system resources & \cite{zhang2024sqlfuse} \\ \hline

\centering Knowledge-to-SQL & \centering DELLM & \centering BIRD, Spider & Execution accuracy & Fine-tuning & Data Expert, Schema Augmentation & Uses tailored Data Expert LLM for knowledge enhancement in SQL generation & Knowledge generation specialized for specific domains only & \cite{hong2024knowledge} \\ \hline


\end{tabular}
}
\end{table*}

\begin{table*}[h]
\centering
\renewcommand{\arraystretch}{1.2}
\caption{State of art methods used in in-context learning for LLM-based text-to-SQL. C1: zero shot and few shots learning, C2: Decomposition, C3: Prompt Optimization, C4: Reasoning Enhancement, C5: Execution Refinement, \cite{zhang2023act}, \cite{liu2024survey}, \cite{fan2024metasql}, \cite{xia2024sql}, \cite{hong2024knowledge}, \cite{gu2024middleware}, \cite{zhang2024structure}, \cite{zhang2024finsql}, \cite{gao2023text}, \cite{vougiouklis2023fastrat}, \cite{dong2023c3}, \cite{fu2023catsql}, \cite{wei2022chain}, \cite{yao2024tree}, \cite{pourreza2024din}. This table was compiled based on information from previous work \cite{hong2024next} }
\label{ICL}
\small
\begin{tabular}{|p{2cm}|c|c|c|p{6cm}|c|}
\hline
\textbf{Methods} & \textbf{Applied LLM} & \textbf{Database} & \textbf{Metrics} & \textbf{Methods Categories} & \textbf{Released} \\ \hline

ACT-SQL & GPT4 & Spider, BIRD & EX, EM, etc. & Prompt Optimization, Reasoning Enhancement & Oct-23 \\ \hline

Schema-free & GPT4 & BIRD & EX & Reasoning Enhancement, Execution Refinement & Aug-24 \\ \hline

MetaSQL & GPT4 & BIRD & EX, EM & Decomposition & Mar-24 \\ \hline

SQL-CRAFT & GPT4 & Spider, BIRD & EX & Reasoning Enhancement, Execution Refinement & Feb-24 \\ \hline

FUXI & GPT4 & BIRD & EX & Reasoning Enhancement, Execution Refinement & Feb-24 \\ \hline

DELLM & GPT4 & Spider, BIRD & EX, VES & Prompt Optimization, Execution Refinement & Feb-24 \\ \hline

SGU-SQL & GPT4 & Spider, BIRD & EX, EM & Decomposition & Feb-24 \\ \hline

FinSQL & LLamA2 & BULL & EX & Zero shot and few shots learning & Jan-24 \\ \hline

DAIL-SQL & GPT Family & Spider, BIRD & EX, EM, VES & Prompt Optimization & Nov-23 \\ \hline

FastRAT & XLM-RoBERTa & Spider, Cspider & EX, EM & Execution Refinement & Nov-23 \\ \hline

CatSQL & BERT & Spider, WikiSQL & EX & Zero shot and few shots learning, Execution Refinement & Nov-23 \\ \hline

C3 & GPT Family & Spider & EX & Zero shot and few shots learning, Prompt Optimization, Execution Refinement & Jul-23 \\ \hline

CoT & GPT4 & Spider, BIRD & EX, VES & Reasoning Enhancement & May-23 \\ \hline

ToT & GPT4 & OTHER & OTHER & Reasoning Enhancement, Execution Refinement & May-23 \\ \hline

DIN-SQL & CodeX, GPT-4 & Spider, BIRD & EX, EM & Decomposition, Execution Refinement & Apr-23 \\ \hline

\end{tabular}
\end{table*}

\subsubsection{In-context Learning}

In in-context learning, the model generates the SQL query based on a given context without any updates to the model parameters. Instead, the model is guided through accurately constructed prompts. In-context learning includes several categories that optimize how queries are generated and improve accuracy. Mathematically, the SQL query generation task is described as
\vspace{-5pt} \begin{equation} Y = f(Q, S, I ; \theta), \end{equation} 

 \emph{Y} is the generated SQL query, \emph{Q} is the NL question set, \emph{S} represents the database schema, \emph{I} is the intermediate reasoning, and $\theta$ are parameters of the pre-trained model. In in-context learning, this formula highlights that the model's output is determined by the input information $(\emph{Q},\emph{S},\emph{I})$. The pre-trained knowledge embedded in $\theta$, remains fixed during the process. The model's performance depends on how effectively the input prompt is engineered to guide the model towards generating accurate SQL queries. Table~\ref{ICL}, classifies  the most well known methods in five categories. The following illustrates all these categories. 

\begin{itemize}

\item \textbf{Zero-Shot and Few-Shot Learning:} 

In zero-shot learning, the model generates SQL queries without any prior exposure to similar examples. the model relies on its pre-trained knowledge to interpret the input and produce SQL queries. For example, in C3's zero-shot prompting of ChatGPT for text-to-SQL tasks, no fine-tuning is needed \cite{dong2023c3}. Zero-shot learning is most effective when the LLM has been pre-trained on a vast corpus that includes SQL-related content. However, in few-shot learning, the model is given a few examples of input-output pairs, to guide its generation of new SQL queries. An example is FinSQL, where models are given a small set of SQL queries and asked to generalize from these examples \cite{zhang2024finsql}. 

\item \textbf{Decomposition:} 

This technique breaks complex queries into simpler sub-queries. Decomposition methods can involve dividing a challenging NL question into multiple SQL queries that are easier to generate. Decomposition improves the model's ability to handle multi-step or nested SQL queries \cite{pourreza2024din}.

\item \textbf{Prompt Optimization:}

This involves refining the input prompts to achieve better SQL generation. By providing more structured or sample prompts, the model is better guided in generating the appropriate SQL queries. Techniques like prompt design and prompt calibration fall under this category, ensuring the prompts are constructed to maximize the model's understanding. Methods like ACT-SQL use prompt optimization to enhance SQL generation by structuring the prompts more efficiently and linking them directly to the database schema \cite{zhang2023act}. 

\item \textbf{Reasoning Enhancement:} 

This methods, such as Chain of Thought (CoT) and Tree of Thoughts (ToT), guide the model to think step-by-step. These approaches enable the model to solve complex queries by generating intermediate reasoning steps before arriving at the final SQL output \cite{yao2024tree},\cite{wei2022chain}. 

\item \textbf{Execution Refinement:} 

These methods involve iterative improvements to the SQL query. The model generates an initial SQL query, executes it, and then refines it based on the results. This method ensures that the generated query is optimized for performance and correctness, minimizing errors in execution \cite{vougiouklis2023fastrat}. 
\end{itemize}

\subsubsection{Fine-Tuning}

\begin{table*}[]
\centering
\renewcommand{\arraystretch}{1.2}
\caption{State of art methods used in fine-tuning (FT) for LLM-based text-to-SQL. C1: Pre-trained, C2: Decomposition, C3: Data Augmented, C4: Enhanced Architecture \cite{pourreza2024sql}, \cite{zhang2024sqlfuse}, \cite{kou2024cllms}, \cite{pourreza2024dts}, \cite{li2024codes}, \cite{xu2023symbol}. This table was compiled based on information from previous work \cite{hong2024next}}
\label{FT}
\small
\begin{tabular}{|p{2.5cm}|c|c|c|p{5cm}|c|}
\hline
\textbf{Methods} & \textbf{LLM} & \textbf{Database} & \textbf{Metrics} & \textbf{Methods Categories} & \textbf{Released Time} \\ \hline

SQL-GEN & LLamA3 & BIRD & EX & Pre-trained, Data Augmented & Aug-24 \\ \hline

SQLfuse & LLamA2-70B & Spider & EX & Pre-trained & Jul-24 \\ \hline

CLMMs & Deepseek & Spider & EX & Enhanced Architecture & Mar-24 \\ \hline

CodeS & StarCoder & Spider, BIRD & EX, VES & Data Augmented & Feb-24 \\ \hline

DTS-SQL & Mistral & Spider, Spider-SYN & EX, EM & Decomposition & Feb-24 \\ \hline

Symbol-LLM & CodeLLaMA & Spider & EM & Data Augmented & Nov-23 \\ \hline

\end{tabular}
\end{table*}

Fine-tuning involves refining the model's internal parameters, $\theta$, using task-specific datasets. Unlike in-context learning methods, where the model's parameters remain fixed and prompts are the primary mechanism of control, fine-tuning updates the model's parameters based on examples from the target task. This process allows the model to become more specialized in tasks like SQL query generation, improving the ability to translate NL questions into accurate SQL queries over time.  

Mathematically, the outcome of this process is represented as a function $g$: \vspace{-5pt}\begin{equation}\theta' = g(\theta, D)\end{equation} where $\theta$ is the pre-trained model's parameters, \emph{D} is the task-specific dataset (pair of questions and SQL queries) and $\theta'$ represents the updated parameters after fine-tuning \cite{kou2024cllms}.

In fine-tuning, the model learns patterns specific to SQL generation, such as understanding database schema and query syntax. This helps it to perform better at generating SQL queries by modifying its internal parameters based on task data, making the model more specialized and accurate for the SQL task. The new parameters, $\theta'$, improves model's ability to generalize across different databases and queries. Currently, a number of studies have been released exploring an improved fine-tuning method. Table ~\ref{FT} shows categorized well-designed fine-tuning methods. 

\begin{itemize}  

\item \textbf{Pre-trained Methods:} 
Pre-trained methods form the backbone of fine-tuning for text-to-SQL systems by leveraging the general knowledge embedded in LLMs, such as GPT, LLaMA, and T5. These models, trained on diverse textual data, are adapted for SQL query generation by fine-tuning with task-specific data. The fine-tuning process enhances their ability to interpret NL and accurately map it to SQL commands across different domains and database schemas \cite{li2023resdsql}. Examples like SQL-GEN show how pre-trained models are fine-tuned with synthetic data for dialect-specific SQL generation, while systems like RESDSQL \cite{li2023resdsql} fine-tune LLMs on datasets like Spider for complex query handling \cite{pourreza2024sql}. 

\item \textbf{Fine-Tuning Decomposition:}
Fine-tuning decomposition methods aim to enhance the performance of LLM on text-to-SQL tasks by breaking down the complex process of query generation into smaller and manageable sub-tasks. The main idea is to address each sub-task individually, thereby allowing the model to better focus and fine-tune its parameters for specific challenges related to text-to-SQL generation. By decomposing the task into stages like schema linking and query formation, model can be trained to handle these distinct processes more effectively than if it were trained on the entire query generation task all at once \cite{pourreza2024dts}. The typical fine-tuning decomposition process involves:

 \begin{itemize}

\item\textbf{Task Segmentation:} breaking down the text-to-SQL conversion into smaller tasks like schema linking and SQL query generation.

\item\textbf{Sequential Fine-Tuning:} Training the model on these sub-tasks in sequence or in parallel so that each sub-task is learned optimally. 
\end{itemize}

\item \textbf{Data Augmented Methods:} The performance of the model is particularly affected by the quality of the training labels during fine-tuning. Inadequate labeling can be counterproductive and often optimal results are not achieved. Rather, if effective augmentation or high-quality data is present, fine tuning is likely to yield results more than even the best fine tuning strategies implemented in low quality or raw data. In text-to-SQL and other problems data-augmented fine-tuning has progressed greatly, as more efforts now aim at improving the data quality rather than the architecture.
As an example, Symbol-LLM has developed an injection and an infusion phase with a focus on improving the data quality during instruction tuning \cite{xu2023symbol}, \cite{li2024codes}.

\item \textbf{Enhance Architecture:}
The generative pre-trained transformer (GPT) framework employs a decoder-only transformer architecture combined with standard auto-regressive decoding for text generation \cite{hong2024next}. However, recent research on the efficiency of large language models (LLMs) has highlighted a shared challenge: when generating long sequences in the auto-regressive paradigm, the attention mechanism increases latency. This issue is pronounced in LLM-based text-to-SQL systems, where generating SQL queries is slower than traditional language modeling, posing a challenge for developing efficient, localized NL interfaces to databases (NLIDB). To address this, Consistency large language models (CLLMs) has been developed with an enhanced model architecture, providing a solution to reduce latency and speed up SQL query generation\cite{kou2024cllms}.  
\end{itemize}

\subsubsection{RAG-based Text-to-SQL System}

RAG-based text-to-SQL systems integrate dynamic retrieval abilities with generative models to improve SQL query generation \cite{zhao2024chat2data}. These systems can be categorized into 5 categories: Dynamic Retrieval, Knowledge-Enhanced Retrieval, Schema-Augmented prompting, Context-Aware Retrieval, and Robustness Enhancement. 

\begin{itemize} 

\item \textbf{Dynamic Retrieval:}

These systems dynamically fetch schema-related information, such as metadata, table descriptions, or previously used queries, to provide relevant context for SQL generation\cite{henriques2024augmenting}. These systems improve adaptability in zero-shot or few-shot scenarios\cite{guo2023retrieval}. However, computational overhead due to repeated retrieval queries can reduce efficiency in real-time applications. 

\item \textbf{Knowledge-Enhanced Retrieval:}

Methods in this category integrate domain-specific unstructured knowledge with schema-based retrieval to improve SQL generation\cite{biswal2024text2sql}. These systems bridge the gap between schema and query understanding by using external knowledge sources \cite{syrja2024retrieval}.  Also these systems are useful for handling domain-specific terminology and queries with incomplete schema information. However, these systems reliant on the quality and availability of domain specific knowledge that may not always be accessible. 

\item \textbf{Schema-Augmented Prompting:}

Schema-Augmented Prompting use retrieved schema information to precise prompts for LLMs\cite{kothyari2023crush4sql}. These systems improved accuracy for complex SQL like those requiring multi-table joins or nested operations\cite{biswal2024text2sql}. However, prompt construction may become verbose because of token limitations and inefficiencies in LLM inference. 

\item \textbf{Context-Aware Retrieval:}

Context-Aware Retrieval systems focus on retrieving relevant context across multi-turn conversations or interactive sessions to generate accurate SQL queries\cite{chen2024fato}. These systems have a good performance on Handling ambiguities in follow-up queries and maintains continuity across dialogue turns\cite{henriques2024augmenting}. However, Context tracking and retrieval can become expensive as conversations grow longer. 

\item \textbf{Robustness Enhancement:}

Robustness systems retrieve alternate schema interpretation or use synonym mappings to handle ambiguity and adversarial challenges in SQL generation. These systems increase resilience to noisy inputs and ambiguous schema mapping. However, if irrelevant data is included, the potential of having inaccurate SQL is increase. 
\end{itemize}

\subsubsection{Novelty and Advantages of RAG-based Systems}

RAG-based text-to-SQL systems introduce several novel features that distinguish them from in-context learning and fine-tuning-based methods:

\begin{itemize}

\item \textbf{1. Dynamic Contextualization:}

Unlike the static fine-tuned models, RAG-based systems can dynamically adapt to new schemas or  domains by retrieving relevant context. 

\item \textbf{2. Enhanced Generalization:}

The retrieval mechanisms allow these systems to excel in zero-shot or few-shot scenarios, making them more flexible than traditional fine-tuning approaches.

\item \textbf{3. Multi-Domain Support:}

By integrating domain-specific knowledge with schema retrieval, RAG-based systems outperform in-context learning models in specialized applications.

\item \textbf{4. Iterative Refinement:}

The feedback loops for SQL generation and execution improve accuracy over time, a feature that is absent in most fine-tuning and in-context learning methods.
\end{itemize}

\subsubsection{Weaknesses of RAG-Based Systems}

While RAG-based systems offer advantages, they also come with certain weaknesses compares to other approaches: 
\begin{itemize} 

\item \textbf{1. Computational Overhead:}

Retrieval mechanisms increase latency, making these systems less efficient for real-time applications. 

\item \textbf{2. Dependency on Retrieval Quality:}

The effectiveness of RAG-based systems is reliant on the quality and relevance of retrieved data. So, poor retrieval can degrade performance. 

\item \textbf{3. Scalability Issues:}

For databases with large schemas or noisy knowledge sources, the retrieval process can become a bottleneck.
\end{itemize}
In Table \ref{compare3}, this study summarized a comparison of RAG-Based, In-Context Learning, and Fine-Tuning methods for LLM-based text-to-SQL. 

\begin{table*}[]
\centering
\renewcommand{\arraystretch}{1.4}
\caption{Comparison of RAG-Based, In-Context Learning, and Fine-Tuning methods for LLM-based text-to-SQL}
\label{compare3}
\small
\begin{tabular}{|p{2.5cm}|p{4.5cm}|p{4.5cm}|p{4.5cm}|}
\hline
\textbf{Aspect} & \textbf{RAG-Based} & \textbf{\begin{tabular}[c]{@{}l@{}}In-Context \ Learning\end{tabular}} & \textbf{Fine-Tuning} \\ \hline
\textbf{Generalization} & \begin{tabular}[c]{@{}l@{}}Excels in zero-shot and few-shot \\ settings with dynamic retrieval.\end{tabular} & \begin{tabular}[c]{@{}l@{}}Limited without extensive prompt \\ engineering.\end{tabular} & \begin{tabular}[c]{@{}l@{}}Domain-specific; requires retraining \\ for new schemas.\end{tabular} \\ \hline
\textbf{Domain Adaptability} & \begin{tabular}[c]{@{}l@{}}High, due to integration of \\ domain-specific knowledge.\end{tabular} & \begin{tabular}[c]{@{}l@{}}Moderate, depends on prompt \\ design.\end{tabular} & \begin{tabular}[c]{@{}l@{}}Low, requires new datasets for \\ domain adaptation.\end{tabular} \\ \hline
\textbf{Efficiency} & \begin{tabular}[c]{@{}l@{}}Slower due to retrieval \\ latency.\end{tabular} & \begin{tabular}[c]{@{}l@{}}Faster but limited by prompt \\ complexity.\end{tabular} & \begin{tabular}[c]{@{}l@{}}Faster inference but requires \\ time-intensive fine-tuning.\end{tabular} \\ \hline
\textbf{\begin{tabular}[c]{@{}l@{}}Implementation \\ Complexity\end{tabular}} & \begin{tabular}[c]{@{}l@{}}High, due to integration of \\ retrieval mechanisms.\end{tabular} & \begin{tabular}[c]{@{}l@{}}Moderate, requires careful \\ prompt design.\end{tabular} & \begin{tabular}[c]{@{}l@{}}High, involves pretraining and \\ schema-specific tuning.\end{tabular} \\ \hline
\end{tabular}
\end{table*}

\section{Graph RAG in Text-to-SQL Systems, A Promising Solution}  

Graph RAG is an advanced framework that integrates graph-based knowledge representation with retrieval-augmented generation techniques \cite{edge2024local}. Unlike traditional RAG systems that retrieve isolated textual data, Graph RAG builds a structured knowledge graph derived from the source documents. This graph organizes entities, relationships, and contextually relevant responses \cite{wu2024medical}. 

\subsection{Novelty of Using Graph RAG
}

Graph RAG introduces a new approach to retrieval-augmented generation by incorporating graph-based structures to organize and retrieve knowledge. This method introduce solutions for  the limitations of traditional RAG systems by emphasizing the connecting data through graph representations to each other \cite{edge2024local}. 

One of its most innovative aspects is its use of modularity detection algorithms, such as Leiden \cite{traag2019louvain}, to partition knowledge graphs into manageable subgraphs. This modular approach facilitates efficient summarization and parallel processing, which is valuable for handling large-scale datasets. Additionally, Graph RAG employs advanced mechanisms like triple graph construction, where entities, their attributes, and relationships are linked to credible sources and controlled vocabularies \cite{wu2024medical}, \cite{edge2024local}.

This approach ensures that generated SQL queries are not only accurate but also grounded in verifiable evidence. Another novelty is its retrieval framework, which combines top-down and bottom-up retrieval strategies to balance context awareness with retrieval efficiency. By integrating these advanced graph-based techniques, Graph RAG redefines how retrieval and reasoning are performed, setting a new standard for RAG systems \cite{edge2024local}, \cite{jeong2024study}. 

\subsection{Why is RAG a Promising Solution for Current LLM-Based Text-to-SQL Limitations?
}
 Graph RAG addresses many of the persistent challenges faced by current LLM-based Text-to-SQL systems\cite{edge2024local}. One of its core strengths is its ability to understand and model database schemas through graph-based relationships\cite{procko2024graph}. Unlike traditional approaches that rely on flat schema descriptions (refer to traditional, straightforward representations of a database schema, typically listing the names of tables, columns, and their basic attributes in a linear or tabular format),  Graph RAG captures relationships between tables, columns, and entities, enabling precise and efficient schema linking\cite{wu2024medical}. This capability is critical for tackling the complexity of modern databases, particularly in cross-domain scenarios. Moreover, Graph RAG's structured graph connections and retrieval mechanisms  improve its ability to resolve ambiguities in queries, such as synonym discrepancies or incomplete schema descriptions\cite{fera2004rag}. Its modular design allows it to generalize across diverse domains by adapting graph construction processes to specific requirements\cite{jeong2024study}. Furthermore, by synthesizing information across graph communities, Graph RAG can handle complex, multi-faceted queries that often confuse traditional systems. Also, the pre-indexed graph structures reduce computational overhead during retrieval, improving efficiency without compromising accuracy. These features position Graph RAG as a robust and adaptable framework for advancing Text-to-SQL systems beyond their current limitations.

\section{Conclusion}
Graph RAG systems are a new paradigm in Retrieval-Augmented Generation, where the mechanism of graph-based structures enhances the generation of SQL queries from natural language. Unlike traditional RAG methods, Graph RAG integrates knowledge and schema information into interlinked graphs, allowing for accurate schema understanding, multi-hop reasoning, and ambiguity resolution. The structured nature of this approach significantly alleviates many limitations that most current LLM-based Text-to-SQL systems suffer from, such as handling complex queries, schema ambiguity, and domain generalization.

Graph RAG systems have indeed made significant progress both in scalability and efficiency by leveraging modularity, community detection, and graph traversal techniques. Graph RAG improves SQL query accuracy by combining strengths in dynamic retrieval, schema augmentation, and robust reasoning and also ensures adaptability across diverse domains and datasets. Graph RAG is a novel approach that combines domain-specific knowledge with complex query generation. It holds great promise in pushing the state-of-the-art for Text-to-SQL systems.

While Graph RAG has shown great promise, there are a number of further areas of exploration and development:

\textbf{Dynamic Schema Adaptation:} 

Future work should investigate how to support real-time updates to graph structures as database schemas evolve. This is crucial for applications in dynamic environments where schema changes are frequent, such as enterprise data lakes or multi-tenant systems.

\textbf{Integration with Conversational Agents:} 

Graph RAG can be used to make Text-to-SQL solutions more user-friendly for non-technical users through the integration of conversational AI. This integration would allow users to interface with databases in natural language-one that the system would dynamically update based on follow-up questions and context shifts.

\textbf{Optimization of Graph Construction:} 

Current graph construction techniques can be resource-intensive, particularly for large-scale databases. Developing lightweight algorithms for efficient graph partitioning, modularity detection, and summarization will improve the scalability of Graph RAG systems, making them more practical for real-world applications.

\textbf{Cross-lingual support of Graph RAG:} 

Extend to support multi-lingual or cross-lingual queries, making it effective for use in a wide number of global use cases. By incorporating language-specific knowledge and translation mechanisms into the graph, Graph RAG systems can facilitate SQL generation across diverse linguistic contexts.

\textbf{Real-time data handling:} 

The challenges related to data freshness and accuracy can be overcome by real-time graph updates combined with strong retrieval mechanisms. This is particularly important for time-critical applications such as financial reporting or live analytics, which require query results to show the most up-to-date data.

\textbf{Improved Explainability:} 

Future versions of Graph RAG should be improved in terms of explainability of query generation processes. These systems can only earn trust and ensure transparency in their outputs by providing users with clear visualizations of graph traversal and reasoning steps.

Whereas graph RAG promises to do so with unprecedented robustness and scalability in a constantly more complex setting of the most modern databases with dynamic interactions, the work done solves crucial limitations that current systems must resolve with proposals toward their future enhancements and makes graph RAG fully empowered for the revolution to come with interacting through natural language towards structured query generation.

\section{Limitation of the State of the Art }
Despite the advances summarized above, a number of challenges remain. These are summarized below. 

\vspace{-5pt}
\subsection{Scalability and Computational Efficiency}
Enhancing LLM-based text-to-SQL systems for large and complex databases without losing computational efficiency is an important challenge. The processing and generation cost of SQL queries remains high, especially with longer sequences and larger datasets. Future solutions will likely focus on model optimizations, more efficient retrieval and storage mechanisms, and specialized indexing techniques to streamline query generation.

\vspace{-5pt}
\subsection{Dynamic Adaptation to Schema Changes}
Most current systems are inefficient in adapting to dynamic, evolving databases without full retraining. Considering that realistic databases will very often experience schema changes and data expansion, the lack of effective techniques, such as incremental learning and flexible architectures, holds back seamless updating of the LLMs and Knowledge Graphs (KGs), potentially leading to reduced query accuracy after some time, most importantly in relation to changing environments.

\vspace{-5pt}
\subsection{Contextual Accuracy and Disambiguity}
Many LLM-based text-to-SQL systems face challenges in handling complex and ambiguous queries where context is not explicitly given. Improving contextual accuracy requires research into how LLMs use structured information from KGs. Enhancing semantic links between user queries and the database schema is needed, and more advanced semantic parsing and disambiguation techniques will help resolve ambiguity.
\vspace{-5pt}
\subsection{Balancing Retrieval-Augmented Generation (RAG) and Fine-Tuning}
While fine-tuning models for specific domains improves performance, RAG offers a way to dynamically incorporate context with less extensive model retraining. The balance between RAG and fine-tuning is an area to be explored, with potential future systems leveraging the strengths of both approaches to minimize training time while maintaining context-sensitive query generation.
\vspace{-5pt}
\subsection{Ethics, Data Privacy, and Interpretability}
The application of LLMs in critical domains like healthcare, finance, and education raises ethical concerns regarding data privacy and model interpretability. Such systems must be transparent, reliable, and respectful of user privacy. Future work needs to establish clear explainability protocols, safe data handling practices, and transparent AI procedures to build trust in LLM-based text-to-SQL systems.
\vspace{-5pt}
\subsection{Human-in-the-Loop and Interactive Querying}
Integration with human feedback is a key future direction. Human-in-the-loop mechanisms will help users to refine and correct generated queries interactively, enhancing model accuracy and transparency. Improved interactivity will not only help build user trust but also provide enhanced learning and error correction opportunities during SQL generation.

\bibliographystyle{IEEEtran}
\bibliography{cite}
\end{document}